\documentclass{article}

\usepackage{arxiv}
\usepackage{xr}
\usepackage[section]{placeins}
\usepackage[utf8]{inputenc} % allow utf-8 input
\usepackage[T1]{fontenc}  % use 8-bit T1 fonts
\usepackage{hyperref}   % hyperlinks
\usepackage{url}  % simple URL typesetting
\usepackage{booktabs}   % professional-quality tables
\usepackage{amsfonts}   % blackboard math symbols
\usepackage{nicefrac}   % compact symbols for 1/2, etc.
\usepackage{microtype}  % microtypography
\usepackage{lipsum}		% Can be removed after putting your text content
\usepackage{graphicx}
\usepackage{xcolor}
\usepackage{doi}
\usepackage{algorithm}
\usepackage{algpseudocode}
\usepackage{amsmath}
\usepackage{multirow}
\usepackage[backend=biber,style=numeric,citestyle=chem-acs]{biblatex}
\usepackage{amssymb}
\usepackage{nameref}
\DeclareFontFamily{U}{stix2bb}{}
\DeclareFontShape{U}{stix2bb}{m}{n} {<-> stix2-mathbb}{}

\title{Operationalizing Serendipity: Multi-Agent AI Workflows for Enhanced Materials Characterization with Theory-in-the-Loop}

\author{
  \textbf{Lance Yao$^{1}$,
  Suman Samantray$^{1}$,
  Ayana Ghosh$^{2}$,
  Kevin Roccapriore$^{3}$,
  }
  \\[1.1\medskipamount]
  \textbf{
  Libor Kovarik$^{1}$,
  Sarah Allec$^{1}$,
  and Maxim Ziatdinov$^{1,\ast}$}
  \\[2\medskipamount]
  $^{1}$\textit{Physical Sciences Division, Pacific Northwest National Laboratory,}\\
  \textit{Richland, WA 99354, USA} \\[0.5\medskipamount]
  $^{2}$\textit{Computational Sciences and Engineering Division, Oak Ridge National Laboratory,}\\
  \textit{Oak Ridge, TN 37831, USA} \\[0.5\medskipamount]
  $^{3}$\textit{AtomQ, Knoxville, TN 37931, USA} 
  \\[1.2\medskipamount]
  $^{\ast}$\texttt{maxim.ziatdinov@pnnl.gov}
}

\providecommand{\headeright}{}
\providecommand{\undertitle}{}
\providecommand{\shorttitle}{}
\renewcommand{\headeright}{Preprint}
\renewcommand{\undertitle}{Preprint}
\renewcommand{\shorttitle}{\emph{Multi-Agent AI Workflows for Enhanced Materials Characterization}}

\hypersetup{
pdftitle={TBA},
pdfsubject={cs.LG, cond-mat.mtrl-sci},
pdfauthor={},
pdfkeywords={TBA},
}

\begin{document}
\maketitle

\begin{abstract}
The history of science is punctuated by serendipitous discoveries, where unexpected observations, rather than targeted hypotheses, opened new fields of inquiry. While modern autonomous laboratories excel at accelerating hypothesis testing, their optimization for efficiency risks overlooking these crucial, unplanned findings. To address this gap, we introduce SciLink, an open-source, multi-agent artificial intelligence framework designed to operationalize serendipity in materials research by creating a direct, automated link between experimental observation, novelty assessment, and theoretical simulations. The framework employs a hybrid AI strategy where specialized machine learning models perform quantitative analysis of experimental data, while large language models handle higher-level reasoning. These agents autonomously convert raw data from materials characterization techniques into falsifiable scientific claims, which are then quantitatively scored for novelty against the published literature. We demonstrate the framework's versatility across diverse research scenarios, showcasing its application to atomic-resolution and hyperspectral data, its capacity to integrate real-time human expert guidance, and its ability to close the research loop by proposing targeted follow-up experiments. By systematically analyzing all observations and contextualizing them, SciLink provides a practical framework for AI-driven materials research that not only enhances efficiency but also actively cultivates an environment ripe for serendipitous discoveries, thereby bridging the gap between automated experimentation and open-ended scientific exploration.
\end{abstract}

\section{Introduction}
The prevailing paradigm of scientific research assumes a linear progression from hypothesis generation to experimental validation. This approach dictates that scientists must first formulate a specific, testable proposition and then design experiments explicitly to confirm or refute it. However, the history of science shows that many transformative discoveries that provide the foundation for modern technology emerged not from targeted inquiry, but from an observation-led process where unexpected findings inspired subsequent theories.

Serendipity-driven breakthroughs in science include Fleming’s discovery of penicillin \cite{aldridge1999discovery}, which revolutionized medicine; Becquerel’s accidental discovery of radioactivity \cite{lodge1912becquerelmemoriallecture}, which changed our understanding of atomic physics; Penzias and Wilson's accidental detection of the cosmic microwave background as persistent antenna noise, which provided foundational evidence for the Big Bang theory \cite{penzias_measurement_1965}; Marshall and Warren’s identification of Helicobacter pylori as the causative agent for ulcers \cite{marshall_unidentified_1984}, which transformed clinical gastroenterology; Plunkett's discovery of Teflon due to a failed refrigerant experiment, which revolutionized industries from aerospace to cookware \cite{plunkett_history_1986}; and Shirakawa, MacDiarmid, and Heeger's discovery of conducting polymers via a catalyst mishap, which paved the way for flexible electronics \cite{shirakawa_synthesis_1977}.  In each case, an investigator, prepared to recognize the significance of an anomaly, turned an unexpected result into a revolutionary scientific advance.

Beyond paradigm-shifting discoveries, the identification of novel behaviors within well-characterized systems - even those tractable by established theoretical frameworks - can be highly consequential for both refining our scientific understanding and enabling new technological applications. This principle is especially potent in materials research. For example, observation of nanometer-sized electron pairing regions in scanning tunneling microscopy experiments provided the missing ingredient for understanding fluctuating superconducting states above a transition temperature in cuprates \cite{gomes_visualizing_2007}. Similarly, observation of polar nanodomains and their ordering behavior not only helped refining understanding of ferroelectricity but also enabled the design of high-performance transducers and novel memory devices based on local electromechanical responses \cite{bonnell_local_2003}.

In recent years, the convergence of robotics, artificial intelligence, and automated analytical techniques has shifted lab experiments from discrete, manual steps to continuous closed-loop processes \cite{darvish_organa_2025, volk_alphaflow_2023, epps_artificial_2020, bran_chemcrow_2023, boiko_autonomous_2023}. As a result, researchers can now conduct autonomous experimentation campaigns that iteratively refine synthesis protocols and characterize materials or biological entities with increased speed and accuracy. These campaigns often begin with computational pre-screening using high-throughput simulations, which have been increasingly automated with the development of workflow management systems (WFMSs) \cite{curtaroloAFLOWAutomaticFramework2012, Jain_Fireworks_2015, Pizzi_AiiDA_2016, Mathew_Atomate_2017, Ganose_Atomate2_2025, Annevelink_AutoMat_2022}, to identify promising candidates, which then proceed to experimental validation using automated or semi-automated platforms. This integrated approach promises to reduce experimental costs and timelines while laying the groundwork for designing novel materials and optimizing complex (bio)chemical processes.

However, the critical challenge is to design and implement AI-driven autonomous systems that do not merely automate the scientific method as it is idealized, but also cultivate an environment ripe for serendipity. This necessitates the development of AI systems that can not only design experiments to test a given hypothesis but also recognize, flag, and investigate data that deviates from existing theories or prevailing wisdom – data that may not even be relevant to the hypothesis that initiated the experiment. Without this capability, the autonomous experimentation, optimized for efficiency, can easily miss important discoveries.

This paper introduces a first step in this direction by introducing a multi-agent AI framework - SciLink - designed to operationalize serendipity and illustrates its application for microscopy and hyperspectral imaging. The process begins with the real-time analysis of raw experimental observations, which it converts into structured scientific claims. The system then interrogates the novelty of these claims against the established body of scientific literature. For any claim flagged as potentially novel, the workflow proceeds to automatically prepare theoretical simulations to provide deeper mechanistic insights and context. By autonomously linking these disparate stages of the scientific process, SciLink aims to accelerate discovery, streamline the research workflow, and, most importantly, provide a robust platform for identifying and investigating unexpected, potentially high-impact scientific findings that might otherwise be overlooked.

\section{Results and Discussion}
\subsection{Overview of SciLink}

\begin{figure}[htbp!]
  \centering
  \includegraphics[width=1.0\textwidth]{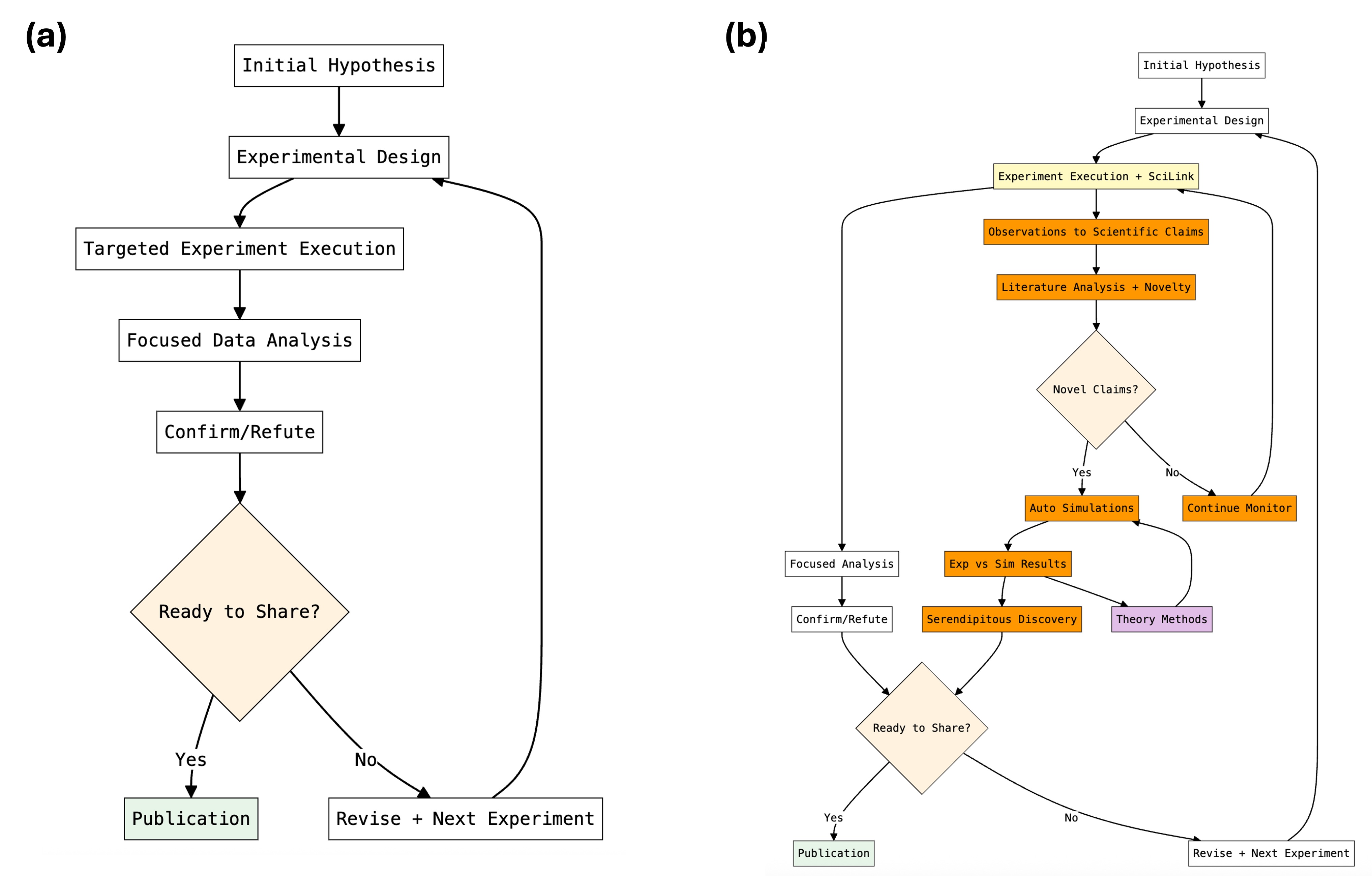}
  \caption{Overall vision comparing traditional and SciLink-enhanced scientific workflows. (a) Traditional hypothesis-driven scientific method following a linear progression from initial hypothesis through experimental design, targeted execution, focused analysis, and validation, ultimately leading to publication or iterative refinement. (b) SciLink-enhanced workflow augments the traditional approach by automatically converting all observations (including those outside the scope of the original inquiry) to scientific claims, performing literature analysis to assess novelty, and triggering automated theoretical simulations. This parallel pathway enables the detection of serendipitous discoveries alongside traditional hypothesis testing.}
  \label{fig:Science_flow}
\end{figure}

Figure ~\ref{fig:Science_flow} illustrates the conceptual shift from the traditional scientific workflow to the SciLink-enhanced approach. The conventional method (Figure~\ref{fig:Science_flow}a) follows a linear, hypothesis-driven path where experiments are designed to test a specific proposition, and data analysis is narrowly focused on confirming or refuting this initial hypothesis. In contrast, the SciLink framework (Figure~\ref{fig:Science_flow}b) augments this process with a parallel, observation-driven pathway designed to operationalize serendipity. Here, all experimental observations, including those incidental to the primary research question, are automatically converted into structured scientific claims. These claims are then systematically evaluated for novelty against the published literature. Potentially novel findings trigger automated theoretical simulations to provide immediate context and theoretical grounding. This dual approach allows SciLink to actively identify, analyze, and investigate unexpected results, thereby creating a fertile ground for serendipitous discoveries that might otherwise be missed.

\begin{figure}[htbp!]
  \centering
  \includegraphics[width=1.0\textwidth]{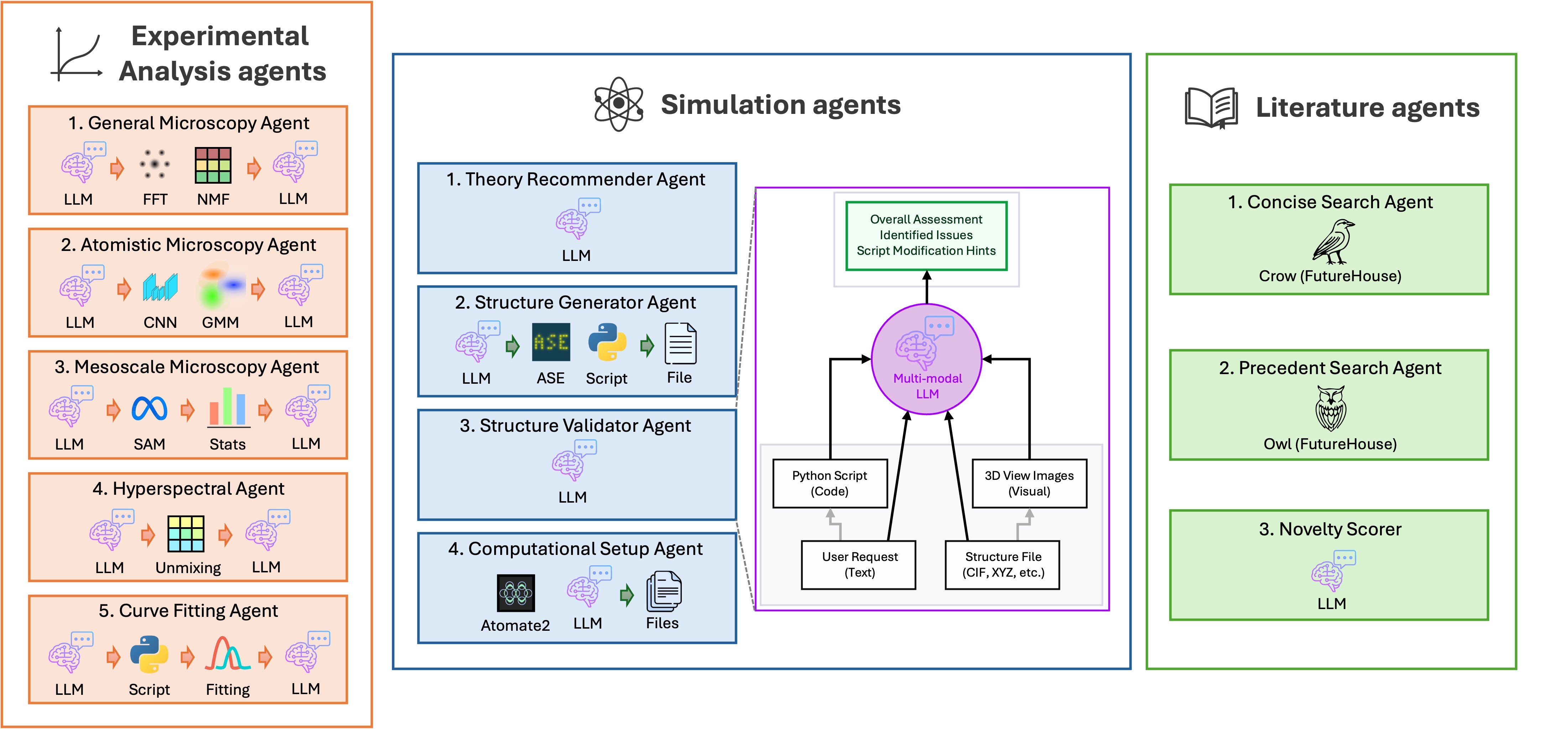}
  \caption{The modular agent categories of the SciLink framework. The architecture comprises three classes of agents: Experimental Analysis agents use a combination of LLMs and specialized techniques (e.g., FFT, CNN, SAM) to analyze diverse data types and generate scientific claims; Simulation agents automate computational modeling, featuring a multi-modal Structure Validator that assesses generated structure by analyzing the user request, script, resulting structure file, and 3D images of the structure; Literature agents provide scientific context by querying external knowledge bases via FutureHouse's scientific agents (Crow and Owl) and using a Novelty Scorer to quantitatively evaluate the originality of the experimental findings.}
  \label{fig:Agents}
\end{figure}

The SciLink framework is built upon three primary categories of autonomous agents, each designed for a specific domain of scientific inquiry: analysis agents that process raw experimental data, such as quantifying features in microscopy images; literature agents that assess scientific novelty by querying publication databases; and simulation agents that streamline the setup of materials science simulations (Figure ~\ref{fig:Agents}). These categories can be organized into different scientific workflows to automate complex research tasks.

The analysis agents utilize a hybrid AI approach, leveraging specialized deep and machine learning models for fast and precise quantitative analysis of domain-specific data, while reserving the Large Language Model (LLM) for higher-level reasoning tasks. This hybrid strategy is orchestrated by an LLM agent in a multi-stage process. Initially, the agent intelligently determines and optimizes parameters for these tools, and executes them to convert raw data into structured, quantitative information, thereby grounding the workflow in high-fidelity analysis. Subsequently, the agent performs its reasoning function by acting on this structured output to interpret results, and generate scientific hypotheses. This approach avoids the inefficiency and potential inaccuracies of forcing a generalist LLM to process raw scientific data \cite{ramachandran_how_2025}, ensuring the optimal AI tool is used for each part of the analysis workflow.

The experimental data analysis toolbox is currently equipped to analyze a wide range of image and hyperspectral data, resolving features from the atomic to the mesoscale. It is applicable to various microscopy and spectroscopy techniques, including scanning transmission electron microscopy, atomic force microscopy, scanning tunneling microscopy, scanning electron microscopy, electron energy loss spectroscopy, and current imaging tunneling spectroscopy. Additionally, the toolbox includes a specialized agent for the analysis of 1D data, such as individual spectra or current-voltage curves. This agent automates the process of fitting the data to physical models by first querying scientific literature for appropriate models, generating the corresponding analysis code, and then interpreting the results.

The literature agents automate scientific novelty assessment through a two-stage process. First, a FutureHouse agent \cite{skarlinski_language_2024} queries existing literature to evaluate a research claim or specific computational parameters. A secondary reasoning agent then analyzes this report against a detailed, structured rubric to assign a quantitative novelty score from 1 (well-established) to 5 (groundbreaking). This scale is designed to reflect common scientific appraisal, where a score of 5 represents a discovery that challenges established theory, and a 4 indicates a high-impact new insight into a specific system. A score of 3 is assigned to partially novel claims where an observation is similar but not identical to published work. Lower scores of 2 and 1 signify that a finding has been previously reported ("scooped") or is considered textbook knowledge, respectively. This quantitative assessment serves as an actionable data point, directly guiding the selection of subsequent theoretical simulations and informing recommendations for the next experiment.

The SciLink framework then extends the role of its analysis agents to next experiment planning. This capability can be triggered after an initial analysis and novelty assessment are complete. The agent synthesizes its own quantitative findings, the generated scientific claims, and the novelty scores from the literature agent to form a comprehensive understanding of the experimental landscape. This synthesized context is then used to prompt the LLM to generate a prioritized list of actionable recommendations for subsequent experiments. Informed by optional access to available experimental capabilities and individual instrument specifications, these recommendations are specific and context-aware, such as suggesting new regions for high-resolution imaging, proposing complementary characterization techniques to probe chemical variations, or recommending in-situ studies to investigate dynamic processes.

Finally, the simulation agents are a suite of specialized modules that automate the translation of scientific concepts into validated, simulation-ready computational models. This suite includes a structure generator agent that converts natural language requests into executable Python scripts leveraging established libraries like the Atomic Simulation Environment (ASE) \cite{Larsen_ASE_2017}. The cornerstone of this group is the multi-modal structure validator agent, which assesses the physical and chemical plausibility of a generated structure by simultaneously analyzing the generating script's logic, the raw atomic coordinates, and rendered 3D images of the structure. This holistic review allows it to identify subtle errors and provide corrective feedback for iterative refinement. Complementing these are a theory recommender agent that prioritizes which novel findings to warrant simulation and a computational setup agent that prepares the final input files, by selecting appropriate parameters based on the scientific objective.

We found that for niche tools and methods in both experimental data analysis and simulations, a best practice is to provide an agent with relevant open source documentation (if available) for in-context learning either simply at the beginning of a request or by using a retrieval-augmented generation system.

These agents serve as modular building blocks for constructing flexible research workflows. In the following sections, we will illustrate how they can be used to perform complex research tasks by autonomously analyzing experimental observations, assessing their novelty against scientific literature, and initiating theoretical simulations.

\begin{figure}[htbp!]
  \centering
  \includegraphics[width=1.0\textwidth]{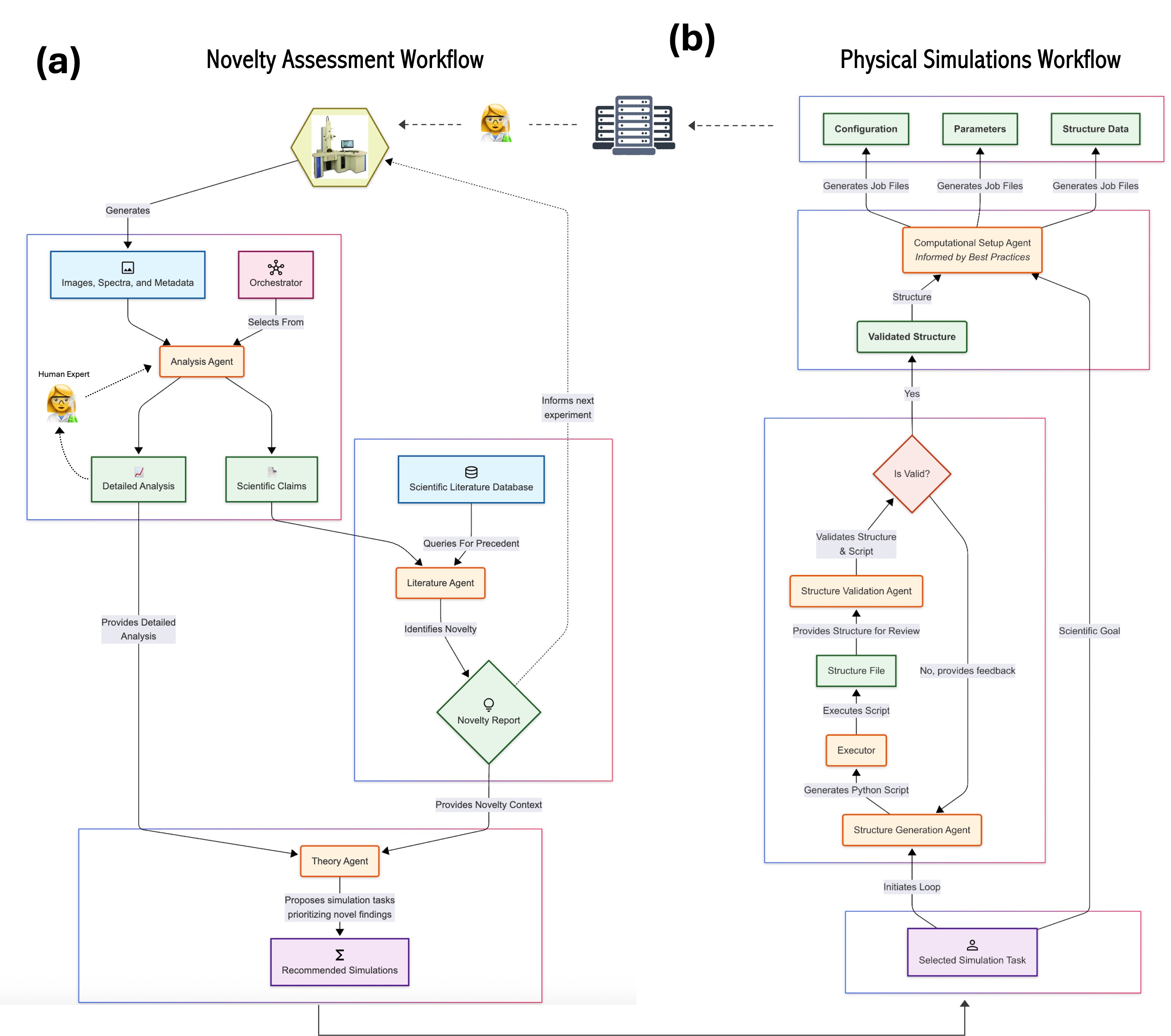}
  \caption{An overview of the major SciLink workflows. The analysis and novelty assessment phase (a) showcases a system for identifying potentially novel phenomena from raw experimental outputs. Initial hypotheses are automatically generated and then systematically vetted for originality against the existing body of scientific literature. This process culminates in a novelty assessment that both informs subsequent theoretical modeling and provides feedback to guide future experiments. The modeling phase (b) demonstrates how a selected novel hypothesis is translated into a computational model. An iterative refinement loop ensures the generated atomic structure is physically and chemically plausible. It then proceeds to generating input files for subsequent DFT calculations, such as INCAR and KPOINTS files for VASP simulations.}
  \label{fig:workflows}
\end{figure}

\subsection{Real-time novelty assessment workflow}

The experiment novelty assessment workflow is designed to automate the critical process of contextualizing new experimental findings within the landscape of existing scientific knowledge (Figure ~\ref{fig:workflows}a). Its primary objective is to bridge the gap between raw data acquisition and novelty assessment, providing researchers with a rapid feedback on the potential impact of their observations. The workflow operates through a modular, multi-stage pipeline that leverages a collection of specialized agents. It begins by ingesting experimental data, such as microscopy images or spectroscopic datasets, and concludes by delivering a structured report that quantitatively scores the novelty of the key scientific claims extracted from that data. This automated approach is engineered to accelerate the research cycle by enabling scientists to efficiently identify and prioritize the most promising results for further investigation.

The architecture of the workflow is composed of two primary stages: (1) Data Analysis and Claim Generation, and (2) Literature Querying and Novelty Scoring. The workflow can also include an optional third stage to generate actionable next steps, which can include one or both of the following: recommendations for theoretical simulations (currently focused on atomistic modeling with density functional theory - DFT), and concrete measurement recommendations to help plan the next experiment. In the initial stage, the workflow employs a data-specific Analyzer module, such as the \textit{MicroscopyAnalyzer} or \textit{SpectroscopyAnalyzer}. This module uses a combination of domain-specific agents to process the raw data and synthesize its findings into a detailed textual analysis. From this analysis, a set of discrete, falsifiable "scientific claims" are formulated. Each claim is a concise statement encapsulating a key observation, such as the identification of a specific crystal defect or a chemical phase distribution.

Once claims are generated, the workflow proceeds to the Literature Querying stage. Each claim is programmatically converted into a natural language research question (e.g., "Has anyone observed a nitrogen vacancy in monolayer tungsten disulfide?"). These questions are then submitted to the \textit{OwlAgent} from FutureHouse \cite{noauthor_futurehouse_2025} designed to provide direct answers and supporting evidence from a vast corpus of scientific literature. In the novelty scoring stage, the response from the literature agent for each claim is passed to a \textit{NoveltyScorer} agent. This agent semantically analyzes the literature report and assign a quantitative novelty score on a scale of 1 (Well-Established) to 5 (Groundbreaking), providing a nuanced evaluation that moves beyond a simple binary determination of novelty. The final output is a comprehensive report detailing the analysis, the claims, the literature evidence, the novelty scores, and any generated recommendations for next steps, such as proposals for computational modeling and suggestions for follow-up experiments, thereby equipping the researcher with a clear and actionable summary of their findings' significance.

\subsection{Physical simulations workflow}

The physical simulation workflow provides an end-to-end automated pipeline for translating high-level, natural language descriptions of material systems into validated, simulation-ready input files for DFT calculations (Figure ~\ref{fig:workflows}b). The workflow is designed to minimize manual setup errors and ensure the physical and chemical soundness of the initial atomic structures through a sophisticated, multi-modal validation and iterative refinement process. This is achieved by orchestrating a series of specialized agents that handle structure generation, validation, and DFT input files creation.

The workflow begins when a user or an agent provides a textual request, such as "a 4x4 graphene supercell with a single vacancy." The \textit{StructureGenerator} agent interprets this request and composes a Python script utilizing the ASE library  to construct the desired atomic model. For security, this AI-generated script is executed within a sandboxed environment to produce an initial structure file. The cornerstone of the workflow is the subsequent multi-modal validation stage, which is performed by the \textit{StructureValidatorAgent}. This agent conducts a comprehensive review of the generated structure by analyzing multiple data streams simultaneously: (1) the logic of the generating Python script, (2) the raw atomic coordinates and lattice vectors within the structure file content, and (3) a series of rendered images of the structure from multiple viewpoints. By cross-referencing these modalities against the original user request, the agent can identify a wide range of potential issues, from subtle deviations in stoichiometry to gross physical inconsistencies like unrealistic bond lengths or atomic clashes.

If the \textit{StructureValidatorAgent} identifies any discrepancies, it generates detailed feedback, including specific, actionable hints on how the generating script should be modified to correct the errors. This feedback initiates an iterative self-correction loop. The \textit{StructureGenerator} receives the validation report and generates a revised script, which is then executed and re-validated. This cycle repeats until the structure passes all validation checks or a predefined number of refinement attempts is reached. 

In contrast to WFMSs that still require domain-expert-prepared input parameters, the workflow then proceeds to generate input files for subsequent DFT calculations - currently targeting the Vienna Ab-initio Simulation Package (VASP) \cite{hafner2008ab-initiosimulationsof} - where the input parameters are selected based on the user's high-level scientific objective and are automatically cross-referenced with published literature to ensure methodological rigor. This step is akin to the recently developed El Agente \cite{Zou_ElAgente_2025} and DREAMS \cite{Wang_DREAMS_2025} frameworks, but we note that the systems of interest here are disordered, defect-containing systems, which introduces additional complexity and requires extensive refinement beyond what is available in published workflows.

\subsection{Use Cases and Examples}

To showcase the versatility and practical application of the SciLink framework, this section presents three use cases from materials characterization, covering both microscopy and hyperspectral spectroscopy. The primary goal of these examples is not to report new scientific discoveries, but rather to demonstrate the framework's operational capabilities and its adaptability to various research challenges.

\begin{figure}[htbp!]
  \centering
  \includegraphics[width=1.0\textwidth]{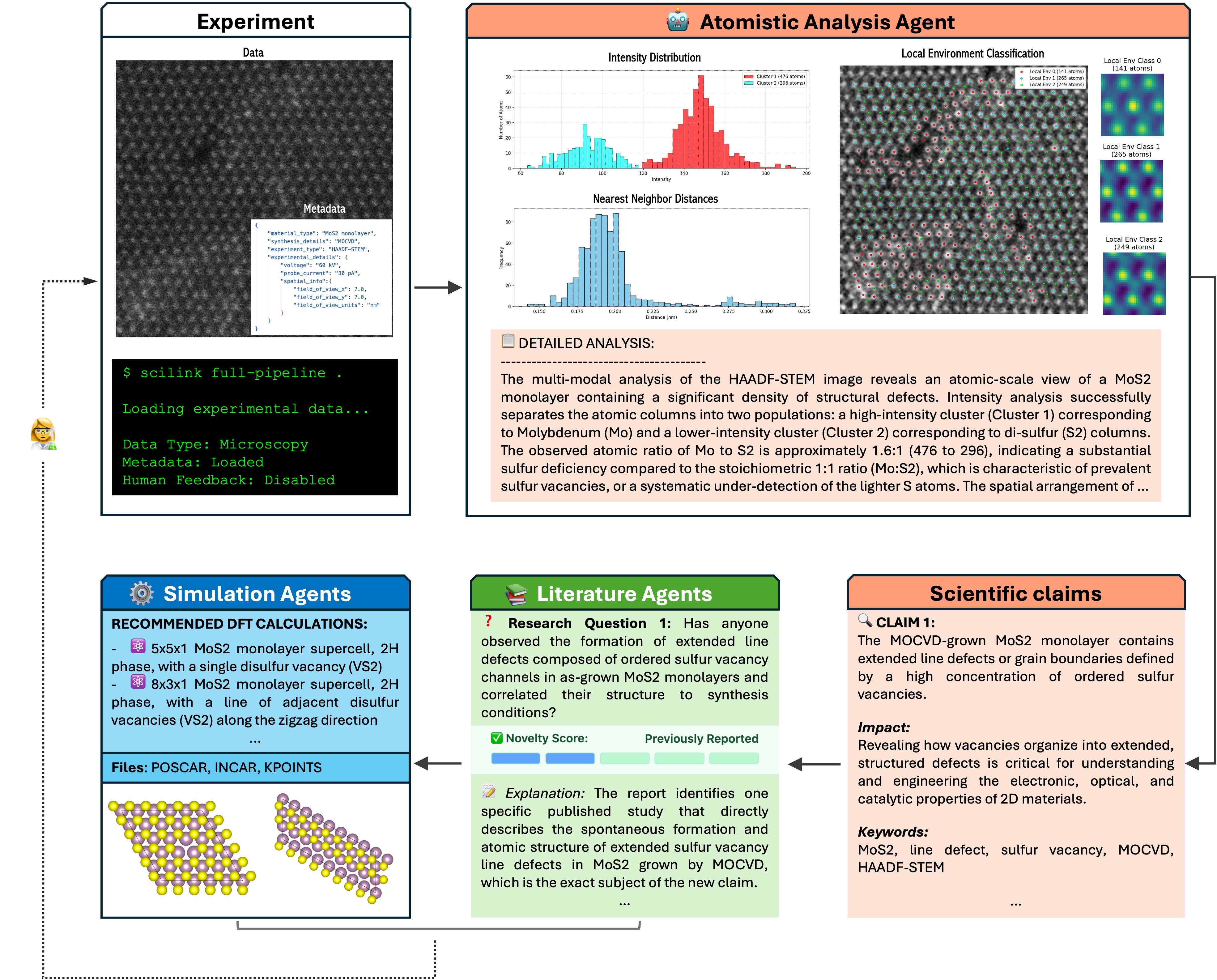}
  \caption{Autonomous analysis and novelty assessment workflow for an atomic-resolved MoS$_2$ monolayer. The workflow is initiated with an experimental HAADF-STEM image and its metadata. The \textit{AtomisticAnalysisAgent} processes the data, producing quantitative outputs: an intensity distribution histogram separating Mo and S atoms, a nearest-neighbor distance plot, and a map of local atomic environments identifying a line defect. Based on this analysis, the agent generates scientific claims, which are converted into research questions for the literature agents. The literature search results are evaluated by a \textit{NoveltyScorer}, which assigns a quantitative score. Finally, the simulation agents uses the analysis to generate validated, simulation-ready DFT inputs for studying the identified defect structure.}
  \label{fig:SciLink_example1}
\end{figure}

\textbf{Example 1: Autonomous Defect Identification in transition metal dichalcogenides.}
To illustrate the Experiment Novelty Assessment workflow, we provided a high-angle annular dark-field scanning transmission electron microscopy (HAADF-STEM) image of of MOCVD-grown molybdenum disulfide (MoS$_2$) monolayer (Figure ~\ref{fig:SciLink_example1}). The workflow orchestrator automatically assigned the \textit{AtomisticAnalysisAgent} based on the quick image inspection and provided metadata. This agent autonomously analyzed the image, using an ensemble of deep convolutional neural networks and a Gaussian mixture model to identify all atomic columns and classify local atomic environments. The agent's key finding was the presence of a high concentration of sulfur vacancies organized into extended line defects. It then formulated this observation into a scientific claim and converted it into a query for the literature agents. The subsequent literature search and \textit{NoveltyScorer} analysis returned a novelty score of 2/5, indicating that this type of ordered vacancy channel in MoS$_2$ is a known (albeit not a textbook) phenomenon. This result highlights the framework's ability to rapidly contextualize experimental findings, preventing researchers from inadvertently pursuing already-published science. Finally, it proceeded to generate a DFT model of the observed structure, enabling a more quantitative validation by calculating properties inaccessible to the imaging technique alone. Despite the finding's low novelty score, this process creates a "digital twin" of the defect, providing a starting data entry that uniquely links the experimental observation to its corresponding theoretical model.

\begin{figure}[htbp!]
  \centering
  \includegraphics[width=1.0\textwidth]{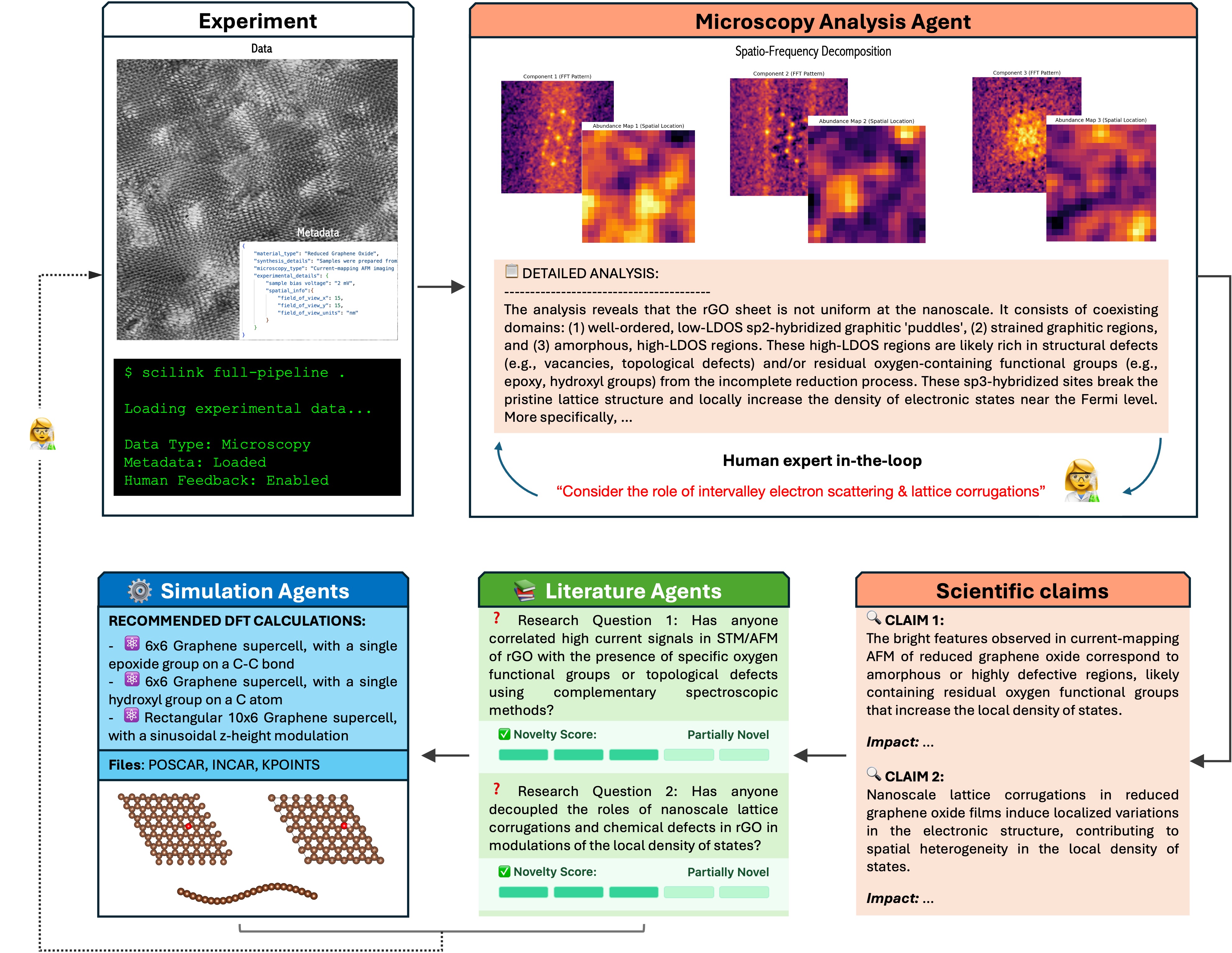}
  \caption{Demonstration of the human-in-the-loop capability within the SciLink workflow for a disordered reduced graphene oxide (rGO) system. An atomic-resolution image of a disordered rGO sheet is processed by the \textit{MicroscopyAnalysisAgent} using spatio-frequency decomposition to identify heterogeneous electronic domains. A human expert provides high-level guidance ("Consider... lattice corrugations"), which is integrated into the agent's reasoning. The workflow generates claims based on both the initial automated analysis and the human-guided path, with both lines of inquiry being assessed as partially novel. The resulting DFT recommendations reflect this hybrid approach, proposing models for both oxygen-related defects and a modulated "wrinkle" structure.}
  \label{fig:SciLink_example2}
\end{figure}

\textbf{Example 2: Human-in-the-Loop Guided Analysis of a Disordered Graphene System.}
This use case demonstrates the hybrid human-AI partnership on a complex, atomic-resolution microscopy image of a highly disordered reduced graphene oxide (rGO) sheet (Figure ~\ref{fig:SciLink_example2}). Due to the significant structural disorder, the framework correctly selected the general \textit{MicroscopyAnalysisAgent}, whose spatio-frequency decomposition method is better suited for identifying patterns in non-crystalline regions. The initial automated analysis revealed heterogeneous electronic domains and led to a claim linking high-current (high-LDOS) regions to amorphous domains containing residual oxygen functional groups or topological defects. At this stage, the human-in-the-loop feature was used, with an expert providing the guidance: "Consider the role of intervalley electron scattering and lattice corrugations." The agent integrated this new context, generating an additional hypothesis that the nanoscale wrinkles themselves induce localized electronic variations. Both the initial AI-autogenerated claim and the subsequent human-guided claim were assessed by the literature agents as partially novel (score 3/5), indicating that the specific nature of these features had not been conclusively established in prior work. This hybrid reasoning was directly reflected in the final DFT recommendations, which included models for both specific oxygen defects (epoxide, hydroxyl) and a "wrinkled" graphene supercell, showcasing how expert intuition can steer the autonomous workflow to explore multiple promising research avenues.

\begin{figure}[htbp!]
  \centering
  \includegraphics[width=1.0\textwidth]{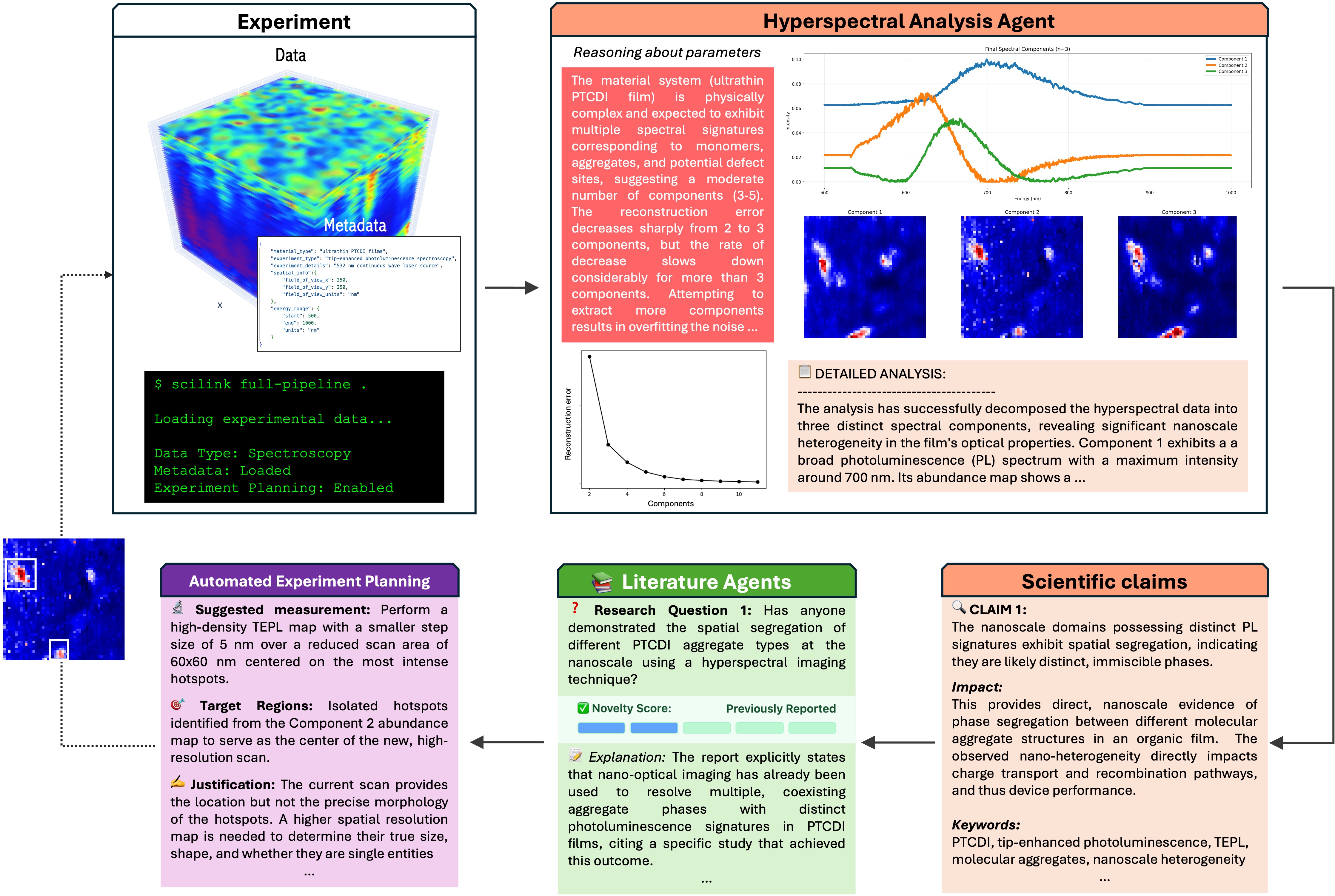}
  \caption{Application of the SciLink framework to hyperspectral data, demonstrating automated next-experiment recommendation. The workflow ingests hyperspectral tip enhanced photoluminescence (TEPL) data from an organic semiconductor thin film. The \textit{HyperspectralAnalysisAgent} performs spectral unmixing to deconvolve the data into constituent spectral signatures and their corresponding 2D spatial abundance maps, revealing nanoscale phase segregation. After claims are generated and assessed against the literature, the workflow leverages this quantitative analysis to propose a targeted follow-up experiment. The agent recommends a higher-spatial-resolution scan focused on the "hotspots" identified in the abundance map of a specific spectral component, providing a precise location and scientific justification for the next measurement.}
  \label{fig:SciLink_example3}
\end{figure}

\textbf{Example 3: Hyperspectral Unmixing and Next-Experiment Recommendation.}
This example showcases the framework's versatility in handling non-image data and its ability to close the experimental loop (Figure ~\ref{fig:SciLink_example3}). The input was a hyperspectral photoluminescence (PL) dataset from an organic semiconductor (PTCDI) thin film. The \textit{HyperspectralAnalysisAgent} applied spectral unmixing to deconvolve the data into three distinct spectral components and their corresponding spatial abundance maps, revealing nanoscale phase segregation and regions with unique excitonic signatures. After generating claims and assessing them against the literature (Novelty Score: 2/5), the workflow's primary output is a recommendation for a follow-up experiment. Based on the hotspots identified in the abundance map of a key spectral component, the system proposed to "Perform a high-density TEPL map with a smaller step size of 5 nm over a reduced scan area... to resolve the precise morphology of the hotspots." This recommendation hypothesizes that even though the macro-scale observation is not novel, the fine-scale morphology and electronic structure within these specific hotspots might be. This demonstrates the framework's capability to analyze spectroscopic data and provide targeted, actionable guidance for the next steps in the experimental research process.

\subsection{Local deployment}

The analysis, literature, and simulation agents are all powered by LLM models. Most frontier LLM models are computationally heavy and are often provided as a cloud service which generates responses to user's input through an API. While SciLink by default utilizes the cloud API Gemini models \cite{comanici2025gemini}, their use may present challenges for scientific applications regarding reproducibility and accessibility, as they can be subject to a service provider's model updates (including abrupt discontinuation) and policy changes. For these reasons, we have also explored the local deployment of SciLink. 

Specifically, SciLink has the option to use Gemma 3 models \cite{team2025gemma} to run experimental data analysis agents locally, including the selection of the analysis tools, the suggestion of the analysis parameters, the interpretation of the results of the analysis, and the generation of the corresponding scientific claims. Gemma 3 is an open source, lightweight series of LLM models from Google designed to work on consumer-grade graphics processing units (GPUs) with the support of both natural language and images as input. Conducting LLM inference on a local machine allows reducing long-term computational costs, ensuring result reproducibility, and provides greater potential for customization. After that, the locally deployed agent still communicates with the cloud by coordinating with the literature and simulation agents to search for novelty and submit simulations. 

Among the different Gemma 3 models, we found that the Quantization-Aware Training (QAT) version of the 27B model delivers performance for interpreting, reasoning, and generating claims that is comparable to cloud-based Gemini models for our specific tasks, while remaining compatible with experimental lab-scale GPUs such as an RTX A6000 with 48GB of graphics memory. On the other hand, the 12B models do not follow the long prompt instruction of SciLink very well, and the non-QAT versions of 27B are too large to fit in the graphic memory. With the rapid development of new LLM models with higher performance and less computational cost, we look forward to the local models as an increasingly viable and practical approach with fast response, enhanced data security, and high fine-tuning flexibility for SciLink.

\begin{figure}[htbp!]
  \centering
  \includegraphics[width=1.0\textwidth]{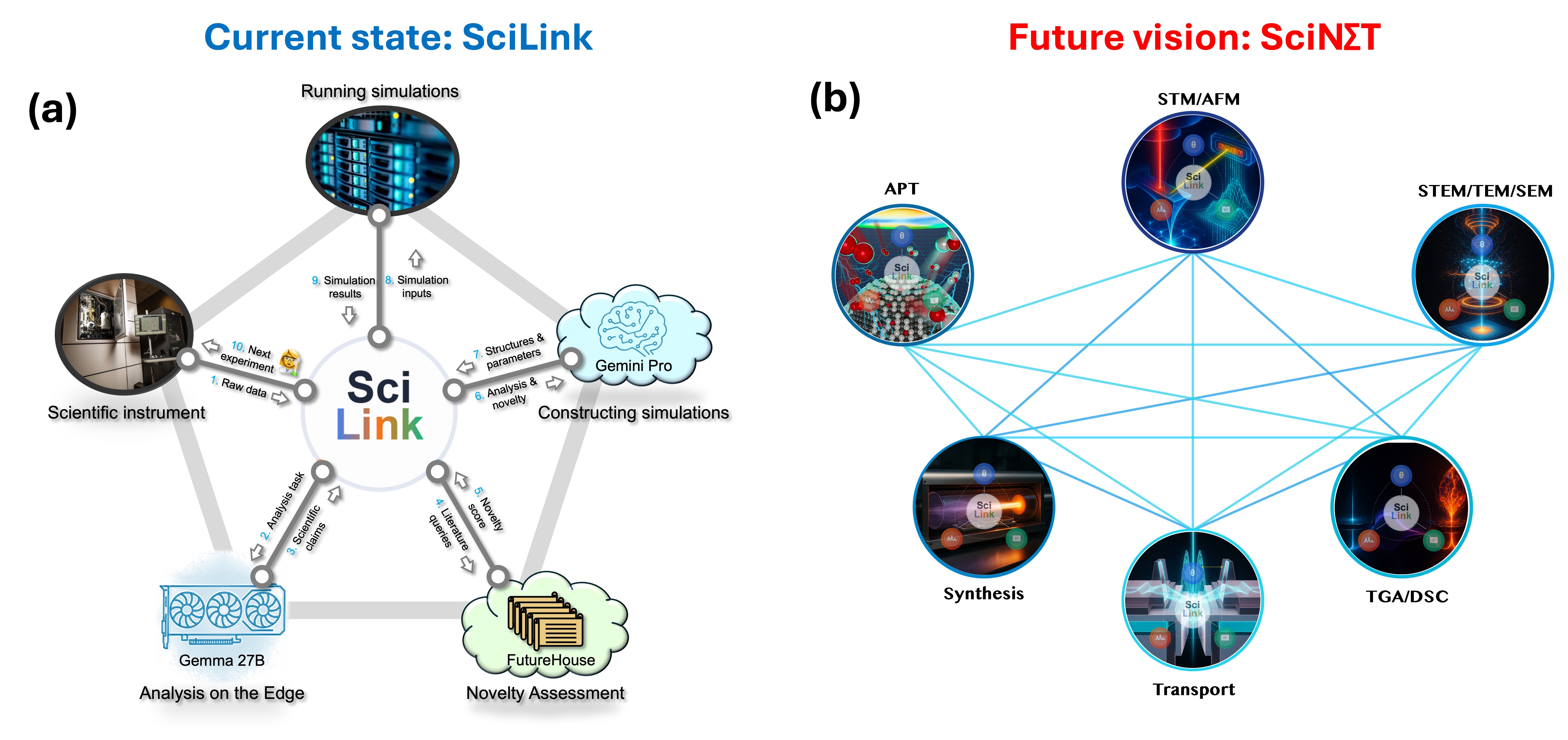}
  \caption{Schematic depiction of the current SciLink framework deployment and the future vision of SciN$\Sigma$T. (a) The current SciLink system creates a closed loop linking a single scientific instrument with on-edge analysis (Gemma 27B), literature-based novelty assessment (FutureHouse), and cloud-powered simulation construction (Gemini Pro). (b) The future vision, SciN$\Sigma$T, depicts a distributed "lab-of-labs" where each node - representing a distinct capability like synthesis or microscopy - is itself a complete SciLink system as shown in (a). This integrated network would enable synergistic ($\Sigma$) research across different, interconnected experimental capabilities.}
  \label{fig:local_future}
\end{figure}

\subsection{Current limitations}

Despite its capabilities, the SciLink framework has several important limitations. The analysis agents, driven by the objective to identify interesting features, can sometimes over-interpret experimental data; the integrated human-in-the-loop feedback mechanism is a crucial feature designed to mitigate this by allowing expert oversight to guide the agent's reasoning. Similarly, the capabilities of the simulation agents are inherently constrained by the underlying libraries they orchestrate, such as the ASE. The novelty assessment also depends on the capabilities of external tools; for instance, as of the moment of writing this paper, the FutureHouse's OWL agent does not analyze figures, meaning it can overlook findings that are not explicitly described in a paper's main text. 

Beyond the specifics of the implementation, a more fundamental challenge is the inherent time discrepancy between theoretical modeling and experimental feedback. While certain experimental processes like sample synthesis can be lengthy, many modern characterization techniques are capable of acquiring key datasets on a timescale of seconds to minutes. In contrast, first-principles approaches like DFT often demand hours to days of intensive computational effort on high-performance computing infrastructures, making it difficult to achieve a truly interactive feedback loop. Although beyond the scope of this work, the integration and development of machine-learned interatomic potentials (MLIPs) \cite{MLIP_Behler_2007, MLIP_Bartók_2018, MLIP_Chan_2019, MLIP_Deringer_2019, MLIP_Zuo_2020, , MLIP_Westermayr_2021, MLIP_MACE_2022, MLIP_Musaelian_2023, MLIP_DeePMD_2023} into this workflow presents a promising pathway to addressing this bottleneck. By leveraging MLIPs, which can offer orders-of-magnitude speedups while preserving \textit{ab initio} accuracy for many materials systems, the workflow could be extended to accelerate high-throughput screening and create more iterative feedback loops between theory and experiment.

Another challenge is the prospective validation of the workflow, as a true test requires a singificant amount of genuinely new, unpublished experimental data. A comprehensive benchmarking of SciLink will therefore necessitate a broader community effort. As such evaluations become available, we are committed to reporting the results in future versions of this work and its corresponding public code repository.

\section{Conclusions and future vision}
In conclusion, we have introduced SciLink, a multi-agent AI framework designed to systematically operationalize serendipity in materials characterization. By creating an automated, closed loop between experimental observation, novelty assessment, and theoretical simulation, SciLink augments the traditional hypothesis-driven scientific method. It ensures that all observations, especially those outside the initial scope of inquiry, are analyzed for their potential scientific impact. Ultimately, this paper's central contribution is the delivery of this methodology as a flexible, open-source tool, designed to be adapted and used for future discovery-focused research.

Our future vision extends SciLink into a comprehensive, networked scientific ecosystem named SciN$\Sigma$T to create a distributed "lab-of-labs," integrating a diverse suite of characterization with automated synthesis platforms. In this interconnected environment, where each node is powered by its own SciLink instance, the multi-agent system will autonomously orchestrate complex, multi-modal experiments across different instruments and laboratories. The goal of SciN$\Sigma$T is to enable a synergistic ($\Sigma$) approach where insights from one technique automatically inform and trigger experiments on another. In such a dynamic, data-rich environment, novel scientific hypotheses can emerge organically, facilitating a holistic understanding of complex materials and accelerating the discovery of new phenomena and functionalities.

\section{Code Availability}
The source code for SciLink is openly available on GitHub at \url{https://github.com/ziatdinovmax/SciLink}.

\section*{Acknowledgments}
This work was supported by the Laboratory Directed Research and Development Program at Pacific Northwest National Laboratory, a multiprogram national laboratory operated by Battelle for the U.S. Department of Energy. Part of the physical simulation workflow development (A.G.) was funded by U.S. Department of Energy, Office of Science, Office of Basic Energy Sciences, Materials Science and Engineering Division.

\printbibliography
\end{document}